\documentclass{article} 
\usepackage{iclr2027_conference,times}

\usepackage{amsmath,amsfonts,bm}

\def\eqref#1{equation~\ref{#1}}

\def\1{\bm{1}}

\DeclareMathAlphabet{\mathsfit}{\encodingdefault}{\sfdefault}{m}{sl}
\SetMathAlphabet{\mathsfit}{bold}{\encodingdefault}{\sfdefault}{bx}{n}

\DeclareMathOperator*{\argmax}{arg\,max}

\usepackage{xurl}
\usepackage{hyperref}
\usepackage{url}
\usepackage{graphicx}
\usepackage{booktabs}
\usepackage[table]{xcolor}
\usepackage{tabularx}
\usepackage{multirow}
\usepackage{caption}
\usepackage{wrapfig}
\usepackage{fvextra}

\usepackage[T1]{fontenc}
\usepackage[utf8]{inputenc}
\usepackage{textcomp}

\usepackage{needspace}
\usepackage[framemethod=default]{mdframed}

\definecolor{rasoFrame}{HTML}{B7CBCD}
\definecolor{rasoTitleBg}{HTML}{EEF4F4}
\definecolor{rasoTitleFg}{HTML}{294D53}

\newmdenv[
  linecolor=rasoFrame,
  linewidth=0.5pt,
  backgroundcolor=white,
  roundcorner=0pt,
  leftmargin=0pt,
  rightmargin=0pt,
  innerleftmargin=9pt,
  innerrightmargin=9pt,
  innertopmargin=7pt,
  innerbottommargin=7pt,
  skipabove=6pt,
  skipbelow=10pt,
  splittopskip=8pt,
  splitbottomskip=6pt,
  frametitlebackgroundcolor=rasoTitleBg,
  frametitlefont=\normalfont\sffamily\bfseries\small\color{rasoTitleFg},
  frametitleaboveskip=5pt,
  frametitlebelowskip=5pt,
  repeatframetitle=true,
  needspace=5\baselineskip,
  nobreak=false,
  startinnercode={\raggedright}
]{rasobox}

\title{Retrieval-Augmented Skill Optimization via Cross-Harness Adaptation}

\author{Jaewon Chu$^1$, 
Ji Soo Lee$^2$, 
Jihwan Park$^2$,
Dohwan Ko$^1$,
Jeehye Na$^2$, 
Seunghun Lee$^2$,\\
\textbf{Taehoon Lee$^2$}, 
\textbf{Minseo Yoon$^2$},
\textbf{Minseok Joo$^1$},
\textbf{Yunyang Xiong$^3$},
\textbf{Hyunwoo J. Kim$^2$\thanks{Corresponding author}}\\
$^1$Korea University, $^2$KAIST, $^3$Meta AI\\
\texttt{allonsy07@korea.ac.kr, hyunwoojkim@kaist.ac.kr}
}

\newcommand{\ie}{\textit{i.e.}}

\iclrfinalcopy 
\begin{document}

\maketitle
\lhead{Retrieval-Augmented Skill Optimization via Cross-Harness Adaptation}

\begin{abstract}
An agent skill is a reusable, actionable natural-language artifact that guides an agent to perform a task effectively under a given harness.
Recent studies have explored the optimization of agent skills, contributing to a growing collection of publicly available skills spanning diverse tasks, domains, and harnesses.
Despite millions of publicly shared skills, existing skill optimization methods largely overlook this accumulated knowledge, instead relying solely on expensive agent rollouts to iteratively refine skills for a target task.
To address this, we propose \textbf{Retrieval-Augmented Skill Optimization (RASO)}, a framework that leverages an external skill corpus as prior knowledge throughout skill optimization.
RASO retrieves relevant knowledge from existing skills and adapts it to the target task and harness via Cross-Harness Adaptation, accounting for mismatches in both domain and harness.
RASO comprises two complementary stages: \textbf{Retrieval-Augmented Skill Initialization (RASI)} constructs a knowledge-grounded initial skill without requiring agent rollouts, while \textbf{Retrieval-Augmented Skill Update (RASU)} iteratively refines the skill by retrieving external knowledge guided by execution feedback.
Across four agent benchmarks and two models, extensive experiments show that RASO consistently outperforms baselines without retrieval-augmented skill initialization and updating.
\end{abstract}
\section{Introduction}
\label{sec:intro}
Large language models (LLMs) are widely deployed as agents within execution harnesses that define available tools, file access, and scoring procedures~\citep{yao2023react, yang2024swe}. 
In these settings, performance depends not only on the parametric knowledge of the underlying model but also on the procedural policies governing task execution~\citep{wang2024voyager, wang2025agent}, commonly referred to as agent skills~\citep{anthropic2025skills, li2026skillsbench}.
An agent skill is a reusable, actionable natural-language artifact that specifies how an agent should accomplish tasks under a given harness. 
Unlike policies encoded in model weights, agent skills are expressed as text, making them readily inspectable, auditable, and transferable across models without modification.

To automatically construct effective agent skills, skill optimization has recently attracted growing attention~\citep{wang2026skillgrad, xia2026grasp, ding2026skillgen, chen2026skillcat}. 
Existing methods iteratively refine skills based on agent experience collected through training-task rollouts under a given harness. 
Through such refinement, millions of skills have been publicly shared~\citep{gitskills}, collectively encoding procedural knowledge accumulated across diverse tasks, models, and harnesses.
Despite this extensive knowledge, existing methods largely rely on the agent's own experience to optimize each skill~\citep{yang2026skillopt, alzubi2026evoskill, ni2026trace2skill, tang2026wikiskill}, requiring costly rollouts for iterative refinement.
As a result, leveraging an external skill corpus as a prior for automatic skill optimization remains largely unexplored.

\begin{figure}[t]
    \centering
    \includegraphics[width=\linewidth]{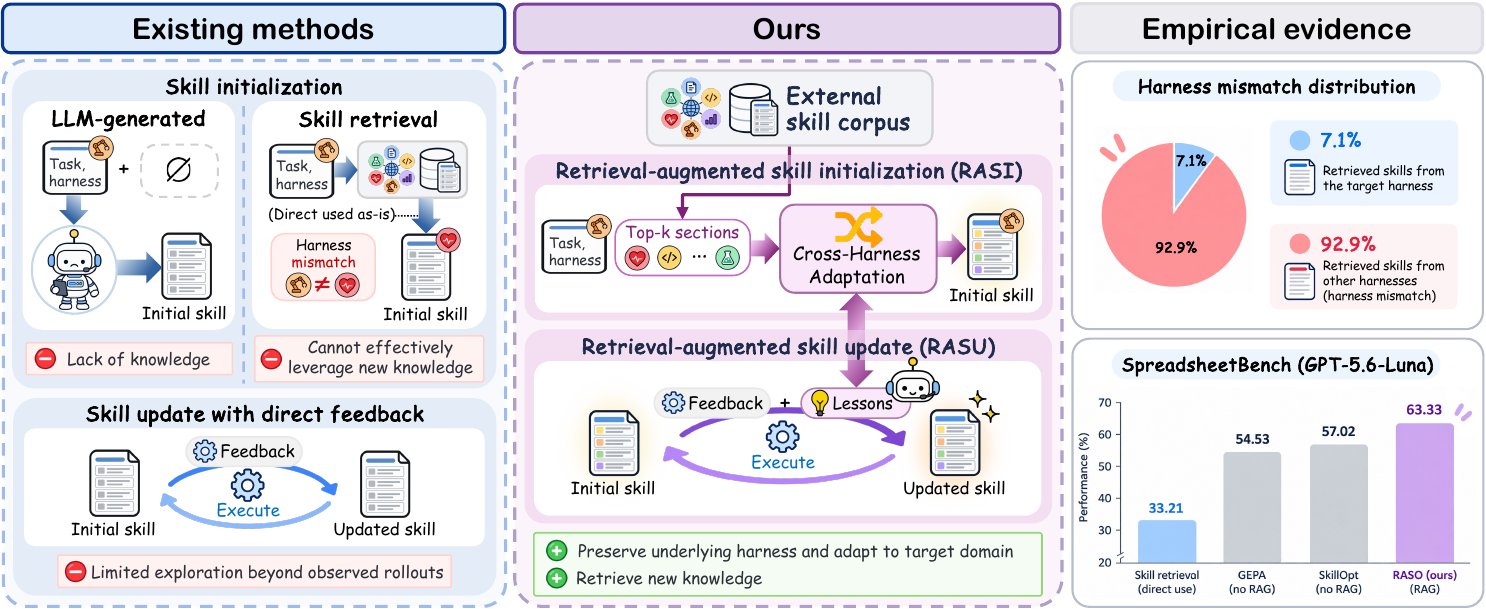}
    \caption{
        \textbf{Comparison of RASO to prior skill optimization.}
        Existing skill optimization methods (left) either generate an initial skill from the LLM alone or directly reuse retrieved skills, then refine them mainly from execution feedback.
        However, retrieved skills often come from mismatched harnesses; on SpreadsheetBench 92.9\% originate from a different harness (right, top). 
        RASO (middle) addresses this by retrieving relevant external skill knowledge and adapting it to the target domain and execution harness through \emph{Cross-Harness Adaptation}, supporting both skill initialization (RASI) and skill update (RASU). 
        Overall, our RASO results in substantially stronger performance than both direct retrieval and retrieval-free optimization methods.
    }
    \label{fig:intro}
    \vspace{-10pt}   
\end{figure}

To this end, we propose \textbf{Retrieval-Augmented Skill Optimization (RASO)}, a skill optimization framework that leverages an external skill corpus as prior knowledge.
As in Fig.~\ref{fig:intro}, RASO comprises two complementary stages, \textbf{Retrieval-Augmented Skill Initialization (RASI)} and \textbf{Retrieval-Augmented Skill Update (RASU)}.
RASI leverages prior knowledge from an external skill corpus to construct an effective initial skill directly from task and harness descriptions, without using any expensive agent rollouts.
Upon observing a failure during task execution, RASU identifies the missing knowledge underlying the failure and retrieves relevant content from the corpus to address it.
However, retrieved skills are originally written for specific tasks and harnesses, potentially encoding domain-specific assumptions or referencing tools unavailable in the target environment.
To address this domain and harness mismatch, RASO introduces Cross-Harness Adaptation, a shared operation employed by both RASI and RASU.
Rather than directly incorporating retrieved content, this operation abstracts away source-domain-specific instruction and re-expresses the underlying procedure in terms of the objects, commands, and units supported by the target harness.

We evaluate RASO through extensive experiments on four agent benchmarks, OfficeQA~\citep{opsahl2026officeqa}, SpreadsheetBench~\citep{ma2024spreadsheetbench}, ALFWorld~\citep{shridhar2021alfworld}, and WebShop~\citep{yao2022webshop}, with two LLMs, Qwen-3.5-9B~\citep{qwen3.5} and GPT-5.6-Luna~\citep{openai2026gpt56}.
Specifically, RASI synthesizes strong initial skills that outperform both retrieved skills and retrieval-free initialization without requiring agent rollouts, providing a strong initialization for subsequent optimization.
Moreover, RASO with RASU consistently outperforms strong skill optimization algorithms, including TextGrad~\citep{yuksekgonul2024textgrad}, GEPA~\citep{agrawal2026gepa}, SkillOpt~\citep{yang2026skillopt}, and WikiSkill~\citep{tang2026wikiskill}.

Our contributions are as follows:
\begin{itemize}
    \item We propose \textbf{Retrieval-Augmented Skill Optimization (RASO)}, a framework that leverages an external skill corpus as prior knowledge for both skill initialization and update. 
    Through its two complementary stages, \textbf{RASI} and \textbf{RASU}, RASO grounds skill construction in retrieved procedural knowledge rather than relying on the optimizer's parametric knowledge.

    \item We introduce Cross-Harness Adaptation, a shared operation that adapts retrieved procedural knowledge to the vocabulary of the target domain and harness. 
    This enables knowledge transfer across diverse domains and harnesses without requiring domain- or harness-matched skills in the corpus.

    \item Across four benchmarks and two models, we demonstrate that RASI improves performance without requiring agent rollouts, while RASU further refines the skill by retrieving the missing knowledge guided by execution feedback. 
    Our analysis further shows that these gains are associated with adapting diverse external knowledge rather than relying on particular source documents.
\end{itemize}


\section{Related Works}

\paragraph{Agent Skills.}
Agent skills are reusable textual documents that provide procedural guidance for accomplishing tasks within an execution harness~\citep{anthropic2025skills}.
Represented as text rather than model parameters, skills are easy to inspect, edit, share, and reuse across models without retraining.
Prior work has made this knowledge explicit through executable skill libraries~\citep{wang2024voyager}, reusable workflows~\citep{wang2025agent}, reflections and insights~\citep{shinn2023reflexion,zhao2024expel}, or reasoning and procedural memories~\citep{ouyang2026reasoningbank,fang2026memp}, while large public collections such as GitSkills~\citep{gitskills} now make this knowledge available at scale.
However, not every skill is useful: its benefit depends on both the procedural knowledge it contains and how well that knowledge matches the target task and harness.
Accordingly, two main approaches have emerged: learning procedural knowledge from the agent's own \emph{execution experience} and reusing knowledge from \emph{external sources}.


\paragraph{Skill Optimization from Execution Experience.}
A major line of work improves agent skill from its own experience, treating the skill as a textual decision variable optimized with execution feedback on the target task.
Methods that optimize prompts or contexts using LLM-generated feedback~\citep{pryzant2023automatic,yang2024large,chu2026agentgrad,zhang2026agentic} are directly applicable to skill optimization, most notably TextGrad~\citep{yuksekgonul2024textgrad}, which backpropagates textual feedback, and GEPA~\citep{agrawal2026gepa}, which evolves prompts by reflecting on execution traces.
Skill-specific optimizers follow the same recipe: SkillOpt~\citep{yang2026skillopt} edits a skill from rollout trajectories and accepts only edits that improve validation performance, WikiSkill~\citep{tang2026wikiskill} compiles agent experience into persistent knowledge, and others distill or refine skills from trajectories~\citep{ni2026trace2skill,chen2026skillcat,moll2026grasp,wang2026skillgrad,alzubi2026evoskill,ding2026skillgen}.
However, these methods primarily optimize skills from observed execution experience, limiting their exploration of procedural knowledge beyond what can be inferred from the agent's own rollouts.
In contrast, our framework supplements execution feedback with procedural knowledge retrieved from an external skill corpus throughout optimization.


\paragraph{Skill Construction from External Knowledge.}
Procedural knowledge that an agent cannot infer from its own experience often already exists: in shared skills, documentation, and the web, motivating a growing line of work that draws on such external knowledge.
Retrieval-based methods, such as SkillRouter~\citep{zheng2026skillrouter}, select relevant skills from large libraries through improved retrievers~\citep{li2026skillflow,miao2026skilllens} or by organizing libraries into graphs and execution structures~\citep{meng2026skillrae,liu2026gos,fu2026se,li2026agentskillos,xia2026grasp}, with dedicated benchmarks for skill retrieval~\citep{cho2026skillret,su2026skill}.
Since retrieved skills may refer to different tasks, tools, or actions, other methods adapt external experience to the target interface~\citep{tang2025agent} or compile external resources into reusable skills ~\citep{pan2026anything2skill,yan2026openskill}.
However, these approaches either require repeated retrieval and adaptation for each task instance during test time or use external knowledge only during skill initialization, without further leveraging it as target-task experience accumulates.
In contrast, we leverage external knowledge throughout the skill optimization process, using it both to construct the initial skill and to further improve the skill during iterative updates.

\section{Problem Formulation}
\label{sec:problem}
We consider a frozen language model $\mathcal{M}$ acting as an agent through an execution harness $h$, which defines the available tools, file access, and observation interface.
A skill $s$ is a natural-language artifact provided as the agent's context to guide task completion under a given harness.
Executing the agent on a task instance $x$ with skill $s$ yields a trajectory $h(\mathcal{M},x,s)$, which is associated with a reward $r(\cdot)\in[0,1]$ by a benchmark-specific evaluator.
When a reference answer is available, the evaluator compares the agent's final output against it, and otherwise uses the environment's native success criterion.
We refer to each agent execution and its corresponding evaluation as a \emph{rollout}.
Given disjoint task splits $\mathcal{D}_\mathrm{train}$, $\mathcal{D}_\mathrm{val}$, and $\mathcal{D}_\mathrm{test}$, candidate skills are constructed from rollouts on $\mathcal{D}_\mathrm{train}$ and selected based on performance on $\mathcal{D}_\mathrm{val}$ as:
\begin{equation}
    s^\star = \argmax_s \mathbb{E}_{x\sim\mathcal{D}_\mathrm{val}}
    \left[r\left(h(\mathcal{M},x,s)\right)\right].
    \label{eq:obj}
\end{equation}
The selected skill $s^\star$ is then evaluated on the held-out $\mathcal{D}_\mathrm{test}$.

Since both $\mathcal{M}$ and $h$ remain fixed throughout optimization, the only decision variable is the natural-language skill $s$.
We include the construction of the initial skill itself in the skill optimization problem, rather than assuming that an initial skill is externally supplied. 
We therefore define skill optimization to encompass both \emph{skill initialization}, which constructs an initial skill from task and harness descriptions without requiring any rollouts, and \emph{skill update}, which refines the skill using results from training rollouts.

\section{RASO: Retrieval-Augmented Skill Optimization}
\label{sec:method}
\begin{figure}[t!]
    \centering
    \includegraphics[width=\linewidth]{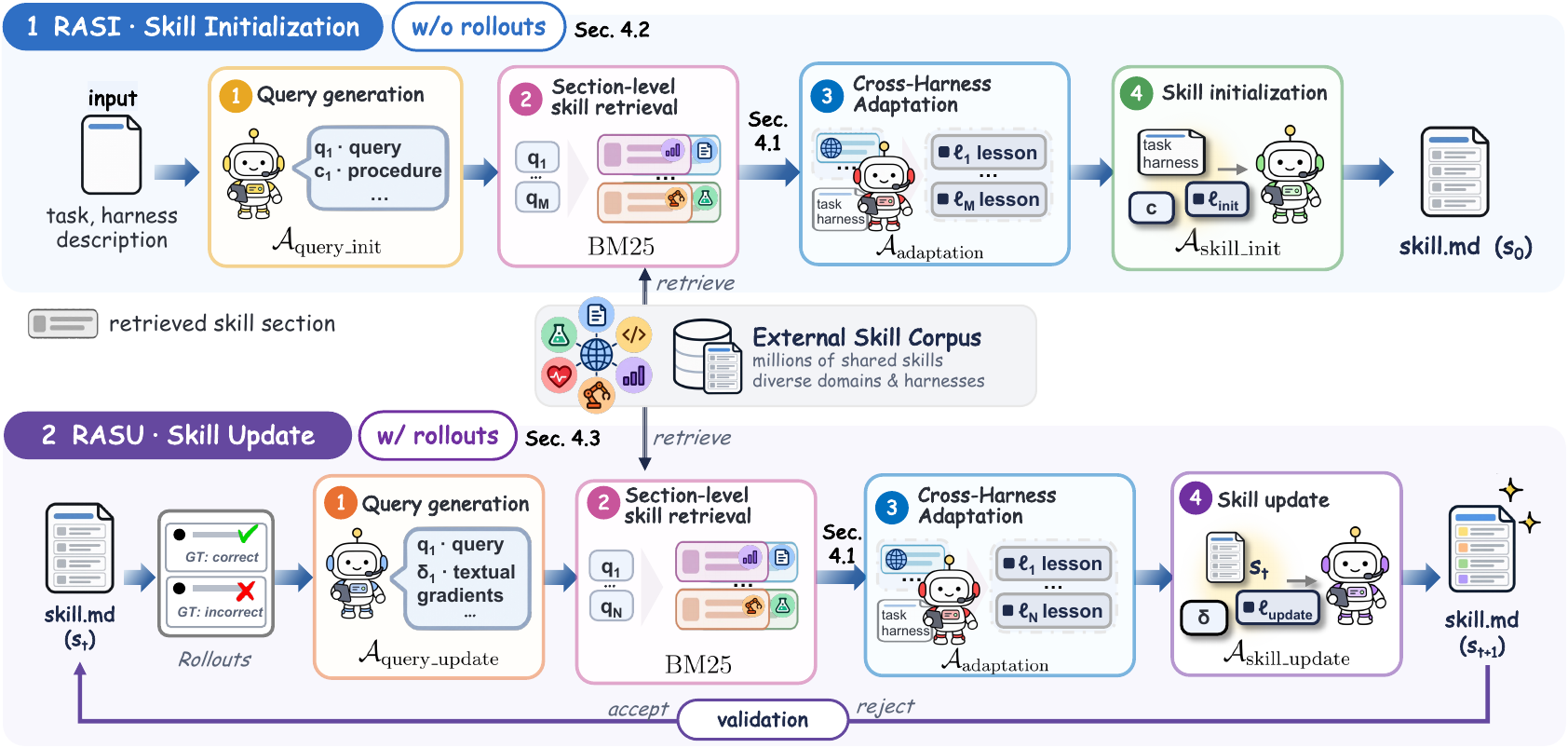}
    \caption{
        \textbf{Overview of RASO.}
        RASI (top, without rollouts) generates retrieval queries from the task and harness descriptions, retrieves the top-$K$ skill sections per query from an external corpus, and adapts them into grounded lessons via Cross-Harness Adaptation to synthesize the initial skill $s_0$.
        RASU (bottom, with rollouts) starts from $s_0$, executes the current skill, and generates retrieval queries from rollout results.
        The retrieved knowledge is adapted into grounded lessons via Cross-Harness Adaptation and incorporated into \texttt{skill.md} to iteratively refine the skill.
        Each subsequent iteration performs new rollouts using the committed skill.
    }
    \label{fig:overview}
\end{figure}

In this section, we introduce \textbf{Retrieval-Augmented Skill Optimization (RASO)}, a framework that evolves agent skills through two complementary stages: skill initialization and skill update, illustrated in Figure~\ref{fig:overview}.
Both stages share a common knowledge retrieval and adaptation mechanism but differ in the rollout evidence available to guide skill optimization.
Section~\ref{main_sec:Retargeting} describes the shared mechanism, which retrieves procedural knowledge from a large-scale skill corpus~\citep{gitskills} and adapts it to the target task and harness through \textit{cross-harness adaptation}.
Section~\ref{main_sec:RASI} introduces \textbf{Retrieval-Augmented Skill Initialization (RASI)}, which constructs an initial skill using only the target task, harness description, and retrieved knowledge, without requiring agent rollouts.
Section~\ref{main_sec:RASU} introduces \textbf{Retrieval-Augmented Skill Update (RASU)}, which leverages execution feedback and retrieved knowledge to iteratively refine the skill.
\subsection{Skill Retrieval and Cross-Harness Adaptation}
\label{main_sec:Retargeting}

We first perform fine-grained, section-level skill retrieval from an external skill corpus to acquire relevant prior knowledge.
We then introduce Cross-Harness Adaptation, a mechanism that bridges the gap between source and target domains and harnesses by transforming retrieved knowledge into actionable guidance tailored to the target task and harness.
Skill retrieval and adaptation constitute a shared pipeline used by both RASI (Section~\ref{main_sec:RASI}) and RASU (Section~\ref{main_sec:RASU}).

\noindent \textbf{Section-level skill retrieval.}
We first divide skill documents into heading-delimited sections to enable fine-grained retrieval.
Since external skill documents are developed for diverse tasks and workflows, retrieving entire documents may introduce irrelevant content and favor documents with similar overall objectives over those containing relevant procedural sections.
Section-level retrieval instead enables us to identify relevant procedural knowledge while improving the signal-to-noise ratio of retrieved content.
Given the resulting section-level corpus $\mathcal{C}$, a query-generation agent formulates a query $q$ and retrieves the top-$K$ most relevant sections using BM25~\citep{robertson2009probabilistic}, \ie, $\mathcal{S}_q = \text{BM25}(q, \mathcal{C}, K)$.

\noindent \textbf{Cross-Harness Adaptation.}
We employ an adaptation agent, denoted by $\mathcal{A}_\mathrm{adaptation}$, to adapt retrieved knowledge to the target task and harness.
Let $c_i$ denote a requirement needing external knowledge to be resolved, such as a specific task procedure in RASI or a textual gradient in RASU.
Given $c_i$, the task description $T$, the harness description $H$, and the corresponding top-$K$ retrieved sections $\mathcal{S}_{q_i}$, the agent produces a concise, actionable lesson $\ell_i$:
\begin{equation}
    \ell_i = \mathcal{A}_\mathrm{adaptation}\left(c_i, T, H, \mathcal{S}_{q_i}\right).
    \label{eq:retargeting}
\end{equation}
Here, the lesson $\ell_i$ serves as a refined knowledge snippet that directly guides the agent to handle the requirement $c_i$ within the target task and harness.

To ensure that each lesson addresses the given requirement and remains valid within the target harness, the adaptation process follows three principles:
(1) remove domain-specific nouns and omit procedures without counterparts in the target harness,
(2) focus exclusively on requirement $c_i$, excluding unrelated issues, and
(3) preserve specific claims about tool or parameter behavior only when corroborated by $H$, prioritizing correctness within the target harness over potentially inaccurate specificity.
Consequently, each lesson is expressed using the objects, commands, and units of the target task and harness.

\subsection{Retrieval-Augmented Skill Initialization (RASI)}
\label{main_sec:RASI}

We introduce \textbf{Retrieval-Augmented Skill Initialization (RASI)}, which constructs an initial skill for a target task under a given harness without requiring agent rollouts.
Performed once at the beginning of skill optimization, RASI comprises four sequential steps: (1) procedure and query generation, (2) section-level skill retrieval, (3) Cross-Harness Adaptation, and (4) skill initialization.

\noindent \textbf{Procedure and query generation.}
Given the task description $T$ and harness description $H$, the query-generation agent $\mathcal{A}_{\mathrm{query\_init}}$ generates a set of requirement-query pairs $(c_i, q_i)$:
\begin{equation}
\{(c_i, q_i)\}_{i=1}^M = \mathcal{A}_{\mathrm{query\_init}}(T, H),
\end{equation}
where $M$ denotes the number of generated pairs. 
Here, each $c_i$ represents a specific task procedure or harness constraint (\textit{e.g.,} multi-turn budget management), and $q_i$ is the retrieval query created to search for external skills that address $c_i$.

\noindent \textbf{Section-level skill retrieval and Cross-Harness Adaptation.}
Given the generated requirement-query pairs, RASI applies the shared retrieval and adaptation pipeline described in Section~\ref{main_sec:Retargeting}.
For each query $q_i$, BM25 retrieves the top-$K$ relevant sections $\mathcal{S}_{q_i}$ from the corpus $\mathcal{C}$.
The adaptation agent $\mathcal{A}_\mathrm{adaptation}$ then transforms these sections into a grounded lesson $\ell_i = \mathcal{A}_\mathrm{adaptation}(c_i, T, H, \mathcal{S}_{q_i})$.
The resulting lessons form $\mathcal{L}_\mathrm{init} = \{\ell_1, \dots, \ell_M\}$, which are used for subsequent skill initialization.

\noindent \textbf{Skill initialization.}
Finally, a skill-initializer agent $\mathcal{A}_\mathrm{skill\_init}$ synthesizes the initial skill $s_0$ from the task description $T$, harness description $H$, identified procedures and requirements $\{c_i\}_{i=1}^M$, and grounded lessons $\mathcal{L}_{\mathrm{init}}$.
The agent integrates each lesson $\ell_i$ into the execution step corresponding to $c_i$, yielding:
\begin{equation}
    s_0 = \mathcal{A}_\mathrm{skill\_init}(T, H, \{c_i\}_{i=1}^M, \mathcal{L}_\mathrm{init}).
\end{equation}
By incorporating retrieved and adapted knowledge before environment interaction, RASI provides a knowledge-grounded initial skill for subsequent optimization without consuming search rollouts.

\subsection{Retrieval-Augmented Skill Update (RASU)}
\label{main_sec:RASU}

Here, we introduce \textbf{Retrieval-Augmented Skill Update (RASU)} that iteratively refines the current skill $s_t$ using execution feedback from agent rollouts.
Complementing the rollout-free initialization of RASI, RASU identifies specific failure modes observed in agent trajectories and retrieves relevant external knowledge to address them.
Given trajectories generated by the execution agent using $s_t$, each RASU iteration comprises four sequential steps: (1) textual gradient and query generation, (2) section-level skill retrieval, (3) Cross-Harness Adaptation, and (4) skill update.

\noindent \textbf{Textual gradient and query generation.}
We first sample a minibatch of tasks from $\mathcal{D}_\mathrm{train}$ and execute agent rollouts using the current skill $s_t$.
A gradient-and-query generator agent $\mathcal{A}_{\mathrm{query\_update}}$ then analyzes the failed trajectories to identify failure mode.
For each failure modes, the agent generates a textual gradient $\delta_i$ based on its parametric knowledge and a targeted retrieval query $q_i$ to acquire relevant external knowledge:
\begin{equation}
\label{eq:query_update}
    \{(q_i, \delta_i)\}_{i=1}^N = \mathcal{A}_{\mathrm{query\_update}}(T, H, s_t, \{\tau_j\}),
\end{equation}
where $N$ denotes the number of generated query and gradient pairs, and $\{\tau_j\}$ denotes the rollout trajectories.

\noindent \textbf{Section-level skill retrieval and Cross-Harness Adaptation.}
Given the failure-driven retrieval queries, RASU applies the shared retrieval and adaptation pipeline described in Section~\ref{main_sec:Retargeting}.
This process yields a set of grounded lessons $\mathcal{L}_{\mathrm{update}} = \{\ell_1, \dots, \ell_N\}$ for subsequent skill update.

\noindent \textbf{Skill update.}
Finally, a skill-updater agent $\mathcal{A}_\mathrm{skill\_update}$ generates a candidate skill $s_{t+1}$ by integrating the current skill $s_t$, textual gradients $\{\delta_i\}_{i=1}^N$, and trajectory-grounded lessons $\mathcal{L}_\mathrm{update}$:
\begin{equation}
    s_{t+1} = \mathcal{A}_\mathrm{skill\_update}(s_t, \{\delta_i\}_{i=1}^N, \mathcal{L}_\mathrm{update}).
\end{equation}
The updater refines the current skill $s_t$ to candidate skill $s_{t+1}$ based on the textual gradients and grounded lessons.
The candidate skill $s_{t+1}$ is accepted only if it outperforms $s_t$ on the validation set $\mathcal{D}_\text{val}$, with $s_t$ retained otherwise.
This pipeline is repeated for a fixed number of iterations, progressively refining the skill through execution feedback and retrieved prior knowledge.

\section{Experiment}
\label{sec:experiment}
\newcolumntype{C}{>{\centering\arraybackslash}X}

\newcommand{\pmse}[1]{{\scriptsize\,$\pm$\,#1}}
\newcommand{\pmseb}[1]{{\scriptsize\,\boldmath$\pm$\,\textbf{#1}}}

\begin{table*}[t]
\centering
\caption{
Agent performance with GPT-5.6-Luna~\citep{openai2026gpt56}
and Qwen-3.5-9B~\citep{qwen3.5}.
We separate skill initialization from skill update;
initialization-only rows report performance at 0 rollouts.
RFSI denotes retrieval-free skill initialization.
Results are mean $\pm$ standard error over three random seeds.
}
\label{tab:skill-performance-v2}

\vspace{6pt}
\begingroup

\definecolor{oursbg}{RGB}{234,242,248}
\definecolor{grpbg}{RGB}{240,240,240}

\setlength{\tabcolsep}{2.5pt}
\renewcommand{\arraystretch}{1.05}
\small

\begin{tabularx}{\textwidth}{
    ll
    *{4}{>{\centering\arraybackslash}X}
}

\toprule




&
& \multicolumn{4}{c}{\textbf{Benchmark}} \\

\cmidrule(lr){3-6}

\textbf{Model}
& \textbf{Method}
& \textbf{OfficeQA}
& \textbf{Spreadsheet}
& \textbf{ALFWorld}
& \textbf{WebShop} \\

\midrule


\rowcolor{grpbg}
\cellcolor{white}
& \multicolumn{5}{l}{\textit{Skill initialization (without rollout)}} \\

\addlinespace[3pt]

& No skill 
& 11.44\pmse{0.70}
& 32.98\pmse{0.12}
& 64.43\pmse{1.38}
& 43.56\pmse{1.01} \\

& SkillRouter 
& 11.44\pmse{0.97}
& 33.21\pmse{0.94}
& 55.97\pmse{1.14}
& 42.20\pmse{0.37} \\

& RFSI 
& 40.11\pmse{0.58}
& 44.40\pmse{2.74}
& 69.40\pmse{1.55}
& 43.89\pmse{1.17} \\

\rowcolor{oursbg}
\cellcolor{white}
& \textbf{RASI} 
& \textbf{45.74}\pmseb{1.40}
& \textbf{49.17}\pmseb{1.52}
& \textbf{72.64}\pmseb{1.63}
& \textbf{45.06}\pmseb{0.40} \\

\cmidrule(lr){2-6}

\rowcolor{grpbg}
\cellcolor{white}
& \multicolumn{5}{l}{\textit{Skill update (with rollout)}} \\

\addlinespace[3pt]

& TextGrad
& 43.80\pmse{1.27}
& 49.64\pmse{3.09}
& 70.65\pmse{0.66}
& 42.32\pmse{3.13} \\

& GEPA
& 43.99\pmse{1.40}
& 54.53\pmse{1.46}
& 71.14\pmse{1.08}
& 45.54\pmse{0.35} \\

& SkillOpt
& 45.54\pmse{2.05}
& 57.02\pmse{3.60}
& 72.64\pmse{0.90}
& 45.17\pmse{0.23} \\

& WikiSkill
& 41.47\pmse{1.18}
& 47.14\pmse{3.98}
& 71.14\pmse{0.66}
& 44.40\pmse{0.64} \\

\rowcolor{oursbg}
\cellcolor{white}
\multirow{-11}{*}[7pt]{\textbf{GPT-5.6-Luna}}
& \textbf{RASO}
& \textbf{49.03}\pmseb{1.85}
& \textbf{63.33}\pmseb{2.69}
& \textbf{74.13}\pmseb{1.00}
& \textbf{46.61}\pmseb{0.11} \\

\midrule


\rowcolor{grpbg}
\cellcolor{white}
& \multicolumn{5}{l}{\textit{Skill initialization (without rollout)}} \\

\addlinespace[3pt]

& No skill 
& 33.14\pmse{1.21}
& 28.81\pmse{0.60}
& 33.09\pmse{0.66}
& 16.27\pmse{0.25} \\

& SkillRouter 
& 34.11\pmse{0.51}
& 23.45\pmse{0.48}
& 31.34\pmse{1.29}
& 9.42\pmse{0.47} \\

& RFSI 
& 34.89\pmse{2.93}
& 27.74\pmse{2.56}
& 42.04\pmse{2.87}
& 12.54\pmse{2.56} \\

\rowcolor{oursbg}
\cellcolor{white}
& \textbf{RASI} 
& \textbf{40.50}\pmseb{0.97}
& \textbf{30.48}\pmseb{0.78}
& \textbf{47.76}\pmseb{3.53}
& \textbf{23.27}\pmseb{0.98} \\

\cmidrule(lr){2-6}

\rowcolor{grpbg}
\cellcolor{white}
& \multicolumn{5}{l}{\textit{Skill update (with rollout)}} \\

\addlinespace[3pt]

& TextGrad
& 36.82\pmse{1.59} & 25.95\pmse{0.63} & 41.04\pmse{2.28} & 10.96\pmse{4.44} \\

& GEPA
& 36.43\pmse{1.97} & 24.05\pmse{1.56} & 43.53\pmse{1.99} & 12.80\pmse{2.20} \\

& SkillOpt
& 37.21\pmse{1.21} & 29.52\pmse{2.21} & 43.03\pmse{3.17} & 13.36\pmse{1.70} \\

& WikiSkill
& 37.40\pmse{3.03} & 29.76\pmse{1.80} & 42.79\pmse{2.93} & 13.43\pmse{2.01} \\

\rowcolor{oursbg}
\cellcolor{white}
\multirow{-11}{*}[7pt]{\textbf{Qwen-3.5-9B}}
& \textbf{RASO}
& \textbf{42.25}\pmseb{1.52}
& \textbf{31.55}\pmseb{0.31}
& \textbf{51.00}\pmseb{2.45}
& \textbf{24.73}\pmseb{0.41} \\

\bottomrule

\end{tabularx}

\endgroup
\end{table*}

We evaluate RASO using two target LLMs: GPT-5.6-Luna~\citep{openai2026gpt56} and Qwen-3.5-9B~\citep{qwen3.5}.
We refer to the model that executes tasks as the target model and the model that generates or updates skill text as the optimizer model.
Unless otherwise specified, we use the same model for both roles across all skill optimization methods.
For skill retrieval, we use GitSkills~\citep{gitskills}, an external skill corpus spanning diverse domains and harnesses.
Our evaluation covers four benchmarks with diverse interaction settings: OfficeQA~\citep{opsahl2026officeqa}, SpreadsheetBench~\citep{ma2024spreadsheetbench}, ALFWorld~\citep{shridhar2021alfworld}, and WebShop~\citep{yao2022webshop}.
For skill initialization, we compare RASI against three strategies: (1) No Skill, where the agent operates without an initialized skill, (2) SkillRouter~\citep{zheng2026skillrouter}, which retrieves a relevant skill from the external corpus, and (3) Retrieval-Free Skill Initialization (RFSI), where the target LLM generates an initial skill directly from the task and harness descriptions without access to the external skill corpus.
For iterative skill optimization, we compare RASO against TextGrad~\citep{yuksekgonul2024textgrad}, GEPA~\citep{agrawal2026gepa}, SkillOpt~\citep{yang2026skillopt}, and WikiSkill~\citep{tang2026wikiskill}.
All experiments are conducted with three random seeds, and we report the mean performance over seeds.

\subsection{Main Results}
\noindent \textbf{Skill initialization.}
We first evaluate the quality of skills produced by different initialization methods, \ie, before any subsequent agent rollout or skill optimization, for both GPT-5.6-Luna and Qwen-3.5-9B in Table~\ref{tab:skill-performance-v2}.
Across both models and all four benchmarks, RASI consistently achieves the highest performance.
With GPT-5.6-Luna, RASI improves over Retrieval-Free Skill Initialization (RFSI) by +5.63 on OfficeQA (45.74 vs. 40.11), +4.77 on Spreadsheet (49.17 vs. 44.40), +3.24 on ALFWorld (72.64 vs. 69.40), and +1.17 on WebShop (45.06 vs. 43.89).
The margin is substantially larger over the No Skill and SkillRouter baselines, particularly on OfficeQA, where both achieve only 11.44.
This advantage also holds for the smaller Qwen-3.5-9B backbone, where RASI outperforms RFSI by +5.61 on OfficeQA, +2.74 on Spreadsheet, +5.72 on ALFWorld, and +10.73 on WebShop.
SkillRouter, which retrieves external skills without Cross-Harness Adaptation, underperforms even the No Skill baseline on several benchmarks, which suggests direct reuse can introduce irrelevant or mismatched procedural knowledge when the retrieved content is not adapted to the target task and harness.
Overall, these results show that combining external knowledge retrieval with LLM-based adaptation, as in RASI, yields a more effective initialization than either retrieval alone or retrieval-free LLM-based skill generation, providing a stronger skill initialization for subsequent optimization.

\paragraph{Skill Update}
We compare RASO with existing skill optimization methods in Table~\ref{tab:skill-performance-v2}.
Following the conventional skill optimization setup, existing methods start from RFSI, where an initial skill is LLM-generated without rollouts or access to an external corpus, and subsequently refine it using their respective rollout-based optimization procedures. 
RASO overall outperforms existing skill optimization methods across both backbones. 
For GPT-5.6-Luna, RASO improves over the strongest competing method by +3.49 on OfficeQA (49.03 vs. 45.54),  +6.31 on SpreadsheetBench (63.33 vs. 57.02), +1.49 on ALFWorld (74.13 vs. 72.64), and +1.07 on WebShop (46.61 vs. 45.54).
Similarly, with Qwen-3.5-9B, RASO improves over the strongest competing method by +4.85 on OfficeQA (42.25 vs. 37.40), +1.79 on SpreadsheetBench (31.55 vs. 29.76), +7.47 on ALFWorld (51.00 vs. 43.53), and +11.30 on WebShop (24.73 vs. 13.43).
These results show that RASO benefits from both retrieval-augmented initialization and subsequent retrieval-augmented skill refinement.
\subsection{Analysis}
\label{sec:analysis}

\begin{table*}[t]
\begingroup
\definecolor{oursbg}{RGB}{234,242,248}
\setlength{\tabcolsep}{4pt}
\renewcommand{\arraystretch}{1}
\begin{minipage}[t]{0.49\textwidth}
\centering
\caption{Ablation on the initialization (Init.) and update components with GPT-5.6-Luna.}
\label{tab:ablation}
\vspace{6pt}
\small
\begin{tabular}{llcc}
\toprule
\multicolumn{2}{c}{\textbf{Method}} & \multicolumn{2}{c}{\textbf{Benchmark}} \\
\cmidrule(lr){1-2}\cmidrule(lr){3-4}
\textbf{Init.} & \textbf{Update} & \textbf{OfficeQA} & \textbf{Spreadsheet} \\
\midrule
RFSI & RFSU & 40.70\pmse{1.67} & 51.67\pmse{2.03} \\
RFSI & \textbf{RASU} & 47.56\pmse{1.85} & 61.07\pmse{2.15} \\
\textbf{RASI} & RFSU & 45.93\pmse{1.58} & 58.45\pmse{1.78} \\
\rowcolor{oursbg} \textbf{RASI} & \textbf{RASU} & \textbf{49.03}\pmseb{1.85} & \textbf{63.33}\pmseb{2.69}\\
\bottomrule
\end{tabular}
\end{minipage}\hfill
\begin{minipage}[t]{0.49\textwidth}
\centering
\caption{Effect of adaptation with GPT-5.6-Luna. RASU used RASI with adaptation as an initial.}
\label{tab:retargeting}
\vspace{6pt}
\small
\begin{tabular}{lccc}
\toprule
& & \multicolumn{2}{c}{\textbf{Benchmark}} \\
\cmidrule(lr){3-4}
\textbf{Method} & \textbf{Adaptation} & \textbf{OfficeQA} & \textbf{Spreadsheet} \\
\midrule
RASI & $\times$ & 41.86\pmse{1.53} & 41.43\pmse{2.83} \\
\rowcolor{oursbg} RASI & $\checkmark$ & \textbf{45.74}\pmseb{1.40} & \textbf{49.17}\pmseb{1.52} \\
\midrule
RASU & $\times$ & 46.70\pmse{2.05} & 57.86\pmse{2.92} \\
\rowcolor{oursbg} \textbf{RASU} & $\checkmark$ & \textbf{49.03}\pmseb{1.85} & \textbf{63.33}\pmseb{2.69}\\
\bottomrule
\end{tabular}
\end{minipage}
\endgroup
\end{table*}

\noindent \textbf{Ablation studies.}
We conduct an ablation study of the RASO components in Table~\ref{tab:ablation}.
Relative to the RFSI + RFSU baseline (retrieval-free settings), RASU improves performance from 40.70 to 47.56 (+6.86 points) on OfficeQA and from 51.67 to 61.07 (+9.40 points) on SpreadsheetBench.
These gains demonstrate the benefit of incorporating retrieved and adapted external knowledge during iterative skill refinement.
Similarly, replacing only the initialization component with RASI improves performance from 40.70 to 45.93 (+5.23 points) on OfficeQA and from 51.67 to 58.45 (+6.78 points) on SpreadsheetBench, demonstrating the benefit of retrieval-augmented initialization under the same subsequent update procedure.
Combining both components leads to the strongest performance, reaching 49.03 on OfficeQA and 63.33 on SpreadsheetBench, corresponding to improvements of +8.33 and +11.66 points over the baseline.
These results indicate that retrieval-augmented initialization and updating provide complementary gains within RASO.

\noindent \textbf{Effect of Cross-Harness Adaptation.}
In Table~\ref{tab:retargeting}, we examine the effect of Cross-Harness Adaptation in both RASI and RASU.
For RASI, applying the adaptation improves performance from 41.86 to 45.74 (+3.88 points) on OfficeQA and from 41.43 to 49.17 (+7.74 points) on SpreadsheetBench.
A similar benefit is observed during skill updating.
Under the same RASI initialization, incorporating the adaptation into RASU improves performance from 46.70 to 49.03 (+2.33 points) on OfficeQA and from 57.86 to 63.33 (+5.47 points) on SpreadsheetBench.
These consistent gains across both stages demonstrate the importance of adapting retrieved knowledge from diverse domains and harnesses to the target task and execution harness for effective skill initialization and iterative refinement.

\begin{table}[t]
\begingroup
\definecolor{oursbg}{RGB}{234,242,248}
\setlength{\tabcolsep}{5.5pt}
\renewcommand{\arraystretch}{1}
\noindent
\begin{minipage}[t]{0.47\textwidth}\vspace{0pt}%
\centering
\caption{Update methods under a fixed RASI initialization with GPT-5.6-Luna. The highlighted row is RASO.}
\label{tab:opt-from-rasi}
\vspace{6pt}
\small
\begin{tabular}{llcc}
\toprule
\multicolumn{2}{c}{\textbf{Method}} & \multicolumn{2}{c}{\textbf{Benchmark}} \\
\cmidrule(lr){1-2}\cmidrule(lr){3-4}
\textbf{Init.} & \textbf{Update} & \textbf{OfficeQA} & \textbf{Spreadsheet} \\
\midrule
 & $-$       & 45.74\pmse{1.40} & 49.17\pmse{1.52} \\
 & TextGrad  & 43.41\pmse{1.97} & 57.97\pmse{3.21} \\
 & GEPA      & 45.74\pmse{1.72} & 58.69\pmse{0.83} \\
 & SkillOpt  & 46.51\pmse{2.10} & 60.24\pmse{1.52} \\
 & WikiSkill & 47.09\pmse{2.42} & 55.00\pmse{4.76} \\
\rowcolor{oursbg} \cellcolor{white}\multirow{-6}{*}{\textbf{RASI}}
 & \textbf{RASU} & \textbf{49.03}\pmseb{1.85} & \textbf{63.33}\pmseb{2.69} \\
\bottomrule
\end{tabular}
\end{minipage}\hfill
\begin{minipage}[t]{0.5\textwidth}\vspace{0pt}%
\centering
\includegraphics[width=\linewidth]{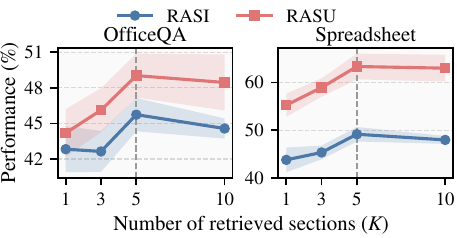}
\captionof{figure}{Effect of the number of retrieved sections $K$ with GPT-5.6-Luna. The dashed line marks our choice of $K=5$, and shading denotes standard error.}
\label{fig:topk}
\end{minipage}%
\endgroup
\end{table}

\noindent \textbf{Comparison with baselines under RASI initialization.}
Table~\ref{tab:opt-from-rasi} compares different skill update methods under the same RASI initialization, thereby isolating the effect of the update procedure.
RASU achieves the highest performance on both benchmarks, reaching 49.03 on OfficeQA and 63.33 on SpreadsheetBench.
Compared with the strongest alternative update method on each benchmark, RASU improves performance by +1.94 points over WikiSkill on OfficeQA (47.09 to 49.03) and by +3.09 points over SkillOpt on SpreadsheetBench (60.24 to 63.33).
Relative to RASI initialization without subsequent updates, RASU yields gains of +3.29 and +14.16 points on OfficeQA and SpreadsheetBench, respectively.
These results show that, under the same retrieval-augmented initialization, RASU provides more effective iterative skill refinement than existing update methods.
\begin{wrapfigure}{r}{0.5\textwidth}
\vspace{-7pt}
\centering
\includegraphics[width=\linewidth]{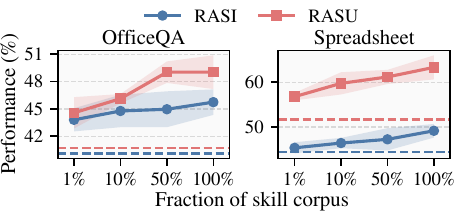}
\vspace{-16pt}
\caption{Effect of skill corpus size with GPT-5.6-Luna. Dashed lines mark 0\% (no corpus), and shading denotes standard error.}
\label{fig:corpus-size}
\vspace{-20pt}
\end{wrapfigure}

\noindent \textbf{Effect of retrieval size.}
Figure~\ref{fig:topk} analyzes the effect of the number of retrieved sections $K$ for both RASI and RASU.
Performance generally improves as $K$ increases from 1 to 5, with the highest performance achieved at $K=5$ on both OfficeQA and SpreadsheetBench.
The gains are particularly pronounced for RASU, indicating that access to multiple relevant sections is useful when refining skills based on rollout feedback.
Further increasing $K$ to 10 slightly degrades performance for both RASI and RASU, suggesting that retrieving additional sections may introduce redundant or less relevant information.
Accordingly, we set $K=5$ for all experiments to balance knowledge coverage against retrieval noise.

\begin{figure}[t!]
    \centering
    \includegraphics[width=\linewidth]{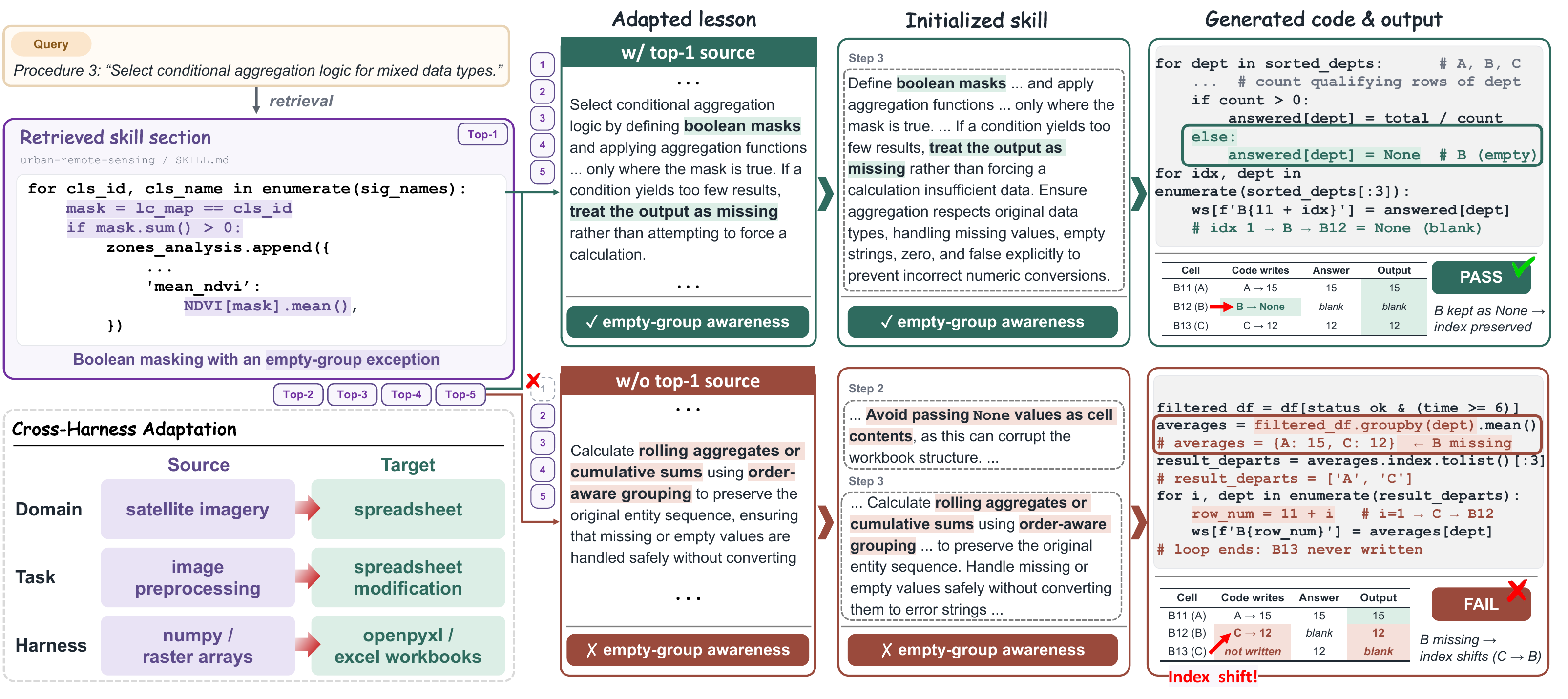}
    \caption{\textbf{Qualitative example of Cross-Harness Adaptation.}
RASI retrieves a skill from a mismatched domain and harness (satellite imagery with numpy/raster arrays ) and adapts its procedure into the target spreadsheet harness. 
We compare the resulting lessons, initial skills, and generated code \textit{with} (top) and \textit{without} (bottom) the top-1 of the five retrieved sections given as input.
Highlights mark the differences between the two runs, which are reflected in the generated code and its output.
Without the top-1 skill, the model overlooks the missing group, causing an index shift.
}

\label{fig:qual} 
    \label{fig:adapataion_qual}
\end{figure}

\noindent \textbf{Effect of skill corpus size.}
Figure~\ref{fig:corpus-size} studies the effect of external skill corpus size for RASI and RASU.
Enabling retrieval from only 1\% of the corpus already yields a clear performance improvement over no-retrieval counterparts (0\%; dashed line), revealing that even a small amount of retrieved external knowledge is beneficial.
As the available corpus grows, performance generally continues to improve, with the largest gains observed on SpreadsheetBench.
Overall, these results suggest that retrieval itself provides an immediate benefit, while increasing corpus size further improves the opportunity to retrieve useful knowledge.

\noindent \textbf{Qualitative results.}
Figure~\ref{fig:qual} presents a SpreadsheetBench task that requires computing the average of qualifying rows for each department. 
For the given data, department B has no qualifying rows; hence, the expected output for its cell is blank.
The top-1 retrieved section comes from an urban remote-sensing skill whose domain and harness differ from the target. Through Cross-Harness Adaptation, RASI rewrites the source's rule of skipping empty groups as ``treat the output as missing,'' and the generated code accordingly leaves B's cell blank.
When this section is removed, the generated code groups only the filtered rows, causing C's mean to shift into B's cell.
This example suggests that cross-harness adaptation transfers the procedural structure and intent of retrieved sections while adapting them to the target harness format.

\section{Conclusion}
\label{sec:conclusion}
We propose Retrieval-Augmented Skill Optimization (RASO), which leverages an external skill corpus as prior knowledge to support both skill initialization and updating.
Rather than directly reusing retrieved skills, RASO applies Cross-Harness Adaptation to transfer relevant procedural knowledge to the target task and execution harness.
RASO combines Retrieval-Augmented Skill Initialization (RASI), which constructs a strong initial skill without agent rollouts, with Retrieval-Augmented Skill Update (RASU), which further refines the skill by retrieving knowledge guided by execution feedback.
Across four agent benchmarks and two model backbones, RASO consistently outperforms retrieval-free skill optimization baselines, demonstrating the value of incorporating external skill knowledge at both initialization and update.





\bibliography{iclr2027_conference}

@article{gitskills,
  title={{GitSkills}: A Dataset of Agent Skills on {GitHub}},
  author={Destefanis, Giuseppe and Graziotin, Daniel and Vaccargiu, Matteo and Ortu, Marco},
  journal={arXiv:2608.10906},
  year={2026}
}

@article{su2026skill,
  title={Skill Retrieval Augmentation for Agentic {AI}},
  author={Su, Weihang and Long, Jianming and Ai, Qingyao and He, Qiaozhi and Tang, Yichen and
          Wang, Changyue and Tu, Yiteng and Wang, Yingbo and Liu, Yiqun},
  journal={arXiv:2604.24594},
  year={2026}
}

@inproceedings{zheng2026skillrouter,
  title={{SkillRouter}: Skill Routing for {LLM} Agents at Scale},
  author={Zheng, Yanzhao and Zhang, Zhentao and Ma, Chao and Yu, Yuanqiang and Zhu, Jihuai and Wu, Yong and
          Xu, Tianze and Dong, Baohua and Zhu, Hangcheng and Huang, Ruohui and Yu, Gang},
  booktitle={COLM},
  year={2026}
}

@article{miao2026skilllens,
  title={{SkillLens}: Adaptive Multi-Granularity Skill Reuse for Cost-Efficient {LLM} Agents},
  author={Miao, Yongliang and Yu, Ziyang and Zhao, Liang and Zhu, Bowen and Haque, Hasibul},
  journal={arXiv:2605.08386},
  year={2026}
}

@article{meng2026skillrae,
  title={{SkillRAE}: Agent Skill-Based Context Compilation for Retrieval-Augmented Execution},
  author={Meng, Xiangcheng and Wang, Shu and Fang, Yixiang},
  journal={arXiv:2605.10114},
  year={2026}
}

@article{fu2026se,
  title={{SE-GoS}: Self-Evolving Graph-of-Skills for Skill Library at Scale},
  author={Fu, Dawei and Jiang, Cheng and Qian, Sitian and Wang, Huainan and Hao, Zhongkai},
  journal={arXiv:2609.08228},
  year={2026}
}

@article{cho2026skillret,
  title={{SkillRet}: A Large-Scale Benchmark for Skill Retrieval in {LLM} Agents},
  author={Kang, Ryangkyung and Cho, Hongcheol and Kim, Youngeun},
  journal={arXiv:2605.05726},
  year={2026}
}

@article{xia2026grasp,
  title={{GraSP}: Graph-Structured Skill Compositions for {LLM} Agents},
  author={Xia, Tianle and Hu, Lingxiang and Sun, Yiding and Xu, Ming and Xu, Lan and Wang, Siying and
          Xu, Wei and Jiang, Jie},
  journal={arXiv:2604.17870},
  year={2026}
}

@article{yang2026skillopt,
  title={{SkillOpt}: Executive Strategy for Self-Evolving Agent Skills},
  author={Yang, Yifan and Gong, Ziyang and Huang, Weiquan and Yang, Qihao and Zhou, Ziwei and Huang, Zisu and
          Li, Yan and Gao, Xuemei and Dai, Qi and Liu, Bei and Qiu, Kai and Yang, Yuqing and
          Chen, Dongdong and Yang, Xue and Luo, Chong},
  journal={arXiv:2605.23904},
  year={2026}
}

@article{tang2026wikiskill,
  title={{WikiSkill}: Compiling Agent Experience into Persistent Knowledge for Skill Evolution},
  author={Tang, Liyan and Rashtchian, Cyrus and Ferng, Chun-Sung and Tomkins, Andrew and Juan, Da-Cheng and
          Vu, Tu},
  journal={arXiv:2608.27454},
  year={2026}
}

@article{ni2026trace2skill,
  title={{Trace2Skill}: Distill Trajectory-Local Lessons into Transferable Agent Skills},
  author={Ni, Jingwei and Liu, Yihao and Liu, Xinpeng and Sun, Yutao and Zhou, Mengyu and Cheng, Pengyu and
          Wang, Dexin and Zhao, Erchao and Jiang, Xiaoxi and Jiang, Guanjun},
  journal={arXiv:2603.25158},
  year={2026}
}

@article{chen2026skillcat,
  title={{SkillCAT}: Contrastive, Assessment-Augmented and Topology-Aware Skill Self-Evolution for {LLM} Agents},
  author={Chen, Kunfeng and Zhong, Qihuang and Liu, Juhua and Du, Bo},
  journal={arXiv:2606.13317},
  year={2026}
}

@article{wang2026skillgrad,
  title={{SkillGrad}: Optimizing Agent Skills Like Gradient Descent},
  author={Wang, Hanyu and Lan, Yifan and Cao, Bochuan and Lin, Lu and Chen, Jinghui},
  journal={arXiv:2605.27760},
  year={2026}
}

@inproceedings{alzubi2026evoskill,
  title={{EvoSkill}: Automated Skill Discovery for Multi-Agent Systems},
  author={Alzubi, Salaheddin and Provenzano, Noah and Bingham, Jaydon and Calvo, Christian Alexander and
          Chen, Weiyuan and Vu, Tu},
  booktitle={COLM},
  year={2026}
}

@inproceedings{ding2026skillgen,
  title={{SkillGen}: Learning Domain Skills for In-Context Sequential Decision Making},
  author={Ding, Ruomeng and Cheng, Wei and Shao, Minglai and Zhao, Chen},
  booktitle={AAAI},
  year={2026}
}

@article{yuksekgonul2024textgrad,
  title={Optimizing generative {AI} by backpropagating language model feedback},
  author={Yuksekgonul, Mert and Bianchi, Federico and Boen, Joseph and Liu, Sheng and Lu, Pan and
          Huang, Zhi and Guestrin, Carlos and Zou, James},
  journal={Nature},
  year={2025}
}

@inproceedings{agrawal2026gepa,
  title={{GEPA}: Reflective Prompt Evolution Can Outperform Reinforcement Learning},
  author={Agrawal, Lakshya A and Tan, Shangyin and Soylu, Dilara and Ziems, Noah and Khare, Rishi and
          Opsahl-Ong, Krista and Singhvi, Arnav and Shandilya, Herumb and Ryan, Michael J and Jiang, Meng and
          Potts, Christopher and Sen, Koushik and Dimakis, Alex and Stoica, Ion and Klein, Dan and
          Zaharia, Matei and Khattab, Omar},
  booktitle={ICLR},
  year={2026}
}

@inproceedings{chu2026agentgrad,
  title={{AgentGrad}: Intervention-guided Prompt Optimization for Multi Agent Systems},
  author={Chu, Jaewon and Seo, Jinwoo and Cho, Jaewon and Na, Jeehye and Xiong, Yunyang and Kim, Youngdae and
          Kim, Hyunwoo J.},
  booktitle={NeurIPS},
  year={2026}
}

@inproceedings{ma2024spreadsheetbench,
  title={{SpreadsheetBench}: Towards Challenging Real World Spreadsheet Manipulation},
  author={Ma, Zeyao and Zhang, Bohan and Zhang, Jing and Yu, Jifan and Zhang, Xiaokang and Zhang, Xiaohan and
          Luo, Sijia and Wang, Xi and Tang, Jie},
  booktitle={NeurIPS},
  year={2024}
}

@article{opsahl2026officeqa,
  title={{OfficeQA Pro}: An Enterprise Benchmark for End-to-End Grounded Reasoning},
  author={Opsahl-Ong, Krista and Singhvi, Arnav and Collins, Jasmine and Zhou, Ivan and Wang, Cindy and
          Baheti, Ashutosh and Oertell, Owen and Portes, Jacob and Havens, Sam and Elsen, Erich and
          Bendersky, Michael and Zaharia, Matei and Chen, Xing},
  journal={arXiv:2603.08655},
  year={2026}
}

@inproceedings{shridhar2021alfworld,
  title={{ALFWorld}: Aligning Text and Embodied Environments for Interactive Learning},
  author={Shridhar, Mohit and Yuan, Xingdi and C{\^o}t{\'e}, Marc-Alexandre and Bisk, Yonatan and
          Trischler, Adam and Hausknecht, Matthew},
  booktitle={ICLR},
  year={2021}
}

@inproceedings{yao2022webshop,
  title={{WebShop}: Towards Scalable Real-World Web Interaction with Grounded Language Agents},
  author={Yao, Shunyu and Chen, Howard and Yang, John and Narasimhan, Karthik},
  booktitle={NeurIPS},
  year={2022}
}

@misc{qwen3.5,
  title  = {{Qwen3.5}: Towards Native Multimodal Agents},
  author = {{Qwen Team}},
  year   = {2026},
  month  = feb,
  url    = {https://qwen.ai/blog?id=qwen3.5}
}

@misc{openai2026gpt56,
  title        = {{GPT-5.6}: Frontier intelligence that scales with your ambition},
  author       = {{OpenAI}},
  year         = {2026},
  month        = jul,
  url = {https://openai.com/index/gpt-5-6/}
}

@inproceedings{liu2026gos,
  title={{Graph-of-Skills}: Dependency-Aware Structural Retrieval for Massive Agent Skills},
  author={Liu, Dawei and Li, Zongxia and Du, Hongyang and Wu, Xiyang and Gui, Shihang and Kuang, Yongbei and
          Sun, Lichao},
  booktitle={EMNLP},
  year={2026}
}

@article{li2026agentskillos,
  title={Organizing, Orchestrating, and Benchmarking Agent Skills at Ecosystem Scale},
  author={Li, Hao and Mu, Chunjiang and Chen, Jianhao and Ren, Siyue and Cui, Zhiyao and Zhang, Yiqun and
          Bai, Lei and Hu, Shuyue},
  journal={arXiv:2603.02176},
  year={2026}
}

@inproceedings{moll2026grasp,
  title={{GRASP}: Gated Regression-Aware Skill Proposer for Self-Improving {LLM} Agents},
  author={Moll, Johannes and Corbeil, Jean-Philippe and Pan, Jiazhen and Hadamitzky, Martin and
          Rueckert, Daniel and Adams, Lisa and Bressem, Keno},
  booktitle={EMNLP},
  year={2026}
}

@inproceedings{pryzant2023automatic,
  title={Automatic Prompt Optimization with ``Gradient Descent'' and Beam Search},
  author={Pryzant, Reid and Iter, Dan and Li, Jerry and Lee, Yin and Zhu, Chenguang and Zeng, Michael},
  booktitle={EMNLP},
  year={2023}
}

@inproceedings{li2026skillflow,
  title={{SkillFlow}: Scalable and Efficient Agent Skill Retrieval System},
  author={Li, Fangzhou and Tagkopoulos, Pagkratios and Tagkopoulos, Ilias},
  booktitle={COLM},
  year={2026}
}

@article{wang2024voyager,
  title={{Voyager}: An Open-Ended Embodied Agent with Large Language Models},
  author={Wang, Guanzhi and Xie, Yuqi and Jiang, Yunfan and Mandlekar, Ajay and Xiao, Chaowei and
          Zhu, Yuke and Fan, Linxi and Anandkumar, Anima},
  journal={TMLR},
  year={2024}
}

@inproceedings{wang2025agent,
  title={Agent Workflow Memory},
  author={Wang, Zora Zhiruo and Mao, Jiayuan and Fried, Daniel and Neubig, Graham},
  booktitle={ICML},
  year={2025}
}

@inproceedings{shinn2023reflexion,
  title={{Reflexion}: language agents with verbal reinforcement learning},
  author={Shinn, Noah and Cassano, Federico and Gopinath, Ashwin and Narasimhan, Karthik R and Yao, Shunyu},
  booktitle={NeurIPS},
  year={2023}
}

@inproceedings{zhao2024expel,
  title={{ExpeL}: {LLM} Agents Are Experiential Learners},
  author={Zhao, Andrew and Huang, Daniel and Xu, Quentin and Lin, Matthieu and Liu, Yong-Jin and
          Huang, Gao},
  booktitle={AAAI},
  year={2024}
}

@inproceedings{yang2024large,
  title={Large Language Models as Optimizers},
  author={Yang, Chengrun and Wang, Xuezhi and Lu, Yifeng and Liu, Hanxiao and Le, Quoc V and Zhou, Denny and
          Chen, Xinyun},
  booktitle={ICLR},
  year={2024}
}

@inproceedings{yao2023react,
    title={{ReAct}: Synergizing reasoning and acting in language models},
    author={Yao, Shunyu and Zhao, Jeffrey and Yu, Dian and Du, Nan and Shafran, Izhak and Narasimhan, Karthik and Cao, Yuan},
    booktitle={ICLR},
    year={2023}
}

@inproceedings{yang2024swe,
  title={{SWE}-agent: Agent-computer interfaces enable automated software engineering},
  author={Yang, John and Jimenez, Carlos and Wettig, Alexander and Lieret, Kilian and Yao, Shunyu and Narasimhan, Karthik and Press, Ofir},
  booktitle={NeurIPS},
  year={2024}
}

@misc{anthropic2025skills,
  title        = {Equipping Agents for the Real World with {Agent Skills}},
  author       = {{Anthropic}},
  year         = {2025},
  month        = oct,
  url = {https://www.anthropic.com/engineering/equipping-agents-for-the-real-world-with-agent-skills}
}

@article{li2026skillsbench,
  title={{SkillsBench}: Benchmarking how well agent skills work across diverse tasks},
  author={Li, Xiangyi and Liu, Yimin and Chen, Wenbo and You, Bingran and Di, Zonglin and He, Yifeng and Zheng, Shenghan and Choe, Kyoung Whan and Sun, Jiankai and Wang, Shuyi and others},
  journal={arXiv:2602.12670},
  year={2026}
}

@article{robertson2009probabilistic,
  title={The probabilistic relevance framework: {BM25} and beyond},
  author={Robertson, Stephen and Zaragoza, Hugo},
  journal={Foundations and Trends in Information Retrieval},
  year={2009},
}

@inproceedings{ouyang2026reasoningbank,
  title={{Reasoningbank}: Scaling agent self-evolving with reasoning memory},
  author={Ouyang, Siru and Yan, Jun and Hsu, I and Chen, Yanfei and Jiang, Ke and Wang, Zifeng and Han, Rujun and Le, Long and Daruki, Samira and Tang, Xiangru and others},
  booktitle={ICLR},
  year={2026}
}

@inproceedings{fang2026memp,
  title={Memp: Exploring agent procedural memory},
  author={Fang, Runnan and Liang, Yuan and Wang, Xiaobin and Wu, Jialong and Qiao, Shuofei and Xie, Pengjun and Huang, Fei and Chen, Huajun and Zhang, Ningyu},
  booktitle={ACL-Findings},
  year={2026}
}

@inproceedings{zhang2026agentic,
  title={Agentic context engineering: Evolving contexts for self-improving language models},
  author={Zhang, Qizheng and Hu, Changran and Upasani, Shubhangi and Ma, Boyuan and Hong, Fenglu and Kamanuru, Vamsidhar and Rainton, Jay and Wu, Chen and Ji, Mengmeng and Li, Hanchen and others},
  booktitle={ICLR},
  year={2026}
}

@article{tang2025agent,
  title={Agent {KB}: Leveraging cross-domain experience for agentic problem solving},
  author={Tang, Xiangru and Qin, Tianrui and Peng, Tianhao and Zhou, Ziyang and Shao, Daniel and Du, Tingting and Wei, Xinming and Xia, Peng and Wu, Fang and Zhu, He and others},
  journal={arXiv:2507.06229},
  year={2025}
}

@article{pan2026anything2skill,
  title={{Anything2Skill}: Compiling External Knowledge into Reusable Skills for Agents},
  author={Pan, Qianjun and Yang, Yutao and Li, Junsong and Zhou, Jie and Chen, Kai and Li, Xin and Chen, Qin and He, Liang},
  journal={arXiv:2606.09316},
  year={2026}
}

@inproceedings{yan2026openskill,
  title={{OpenSkill}: Open-World Self-Evolution for LLM Agents},
  author={Yan, Zhiling and Song, Dingjie and Zhang, Hanrong and Liang, Wei and Zhang, Yuxuan and Dai, Yutong and He, Lifang and Yu, Philip S and Xu, Ran and Li, Xiang and others},
  booktitle={EMNLP},
  year={2026}
}
\bibliographystyle{iclr2027_conference}

\newpage

\appendix

\section{Appendix}

\subsection{Implementation Details}
\label{app:impl}
For all update methods, we run skill optimization for 2 epochs over $\mathcal{D}_{\text{train}}$ with a minibatch size of 40 tasks per iteration.
For Qwen-3.5-9B, we use a sampling temperature of 0.0 for agent rollouts and 0.7 for the optimizer model, and serve the model with vLLM on a single NVIDIA RTX A6000 GPU.
For GPT-5.6-Luna, we set the reasoning effort to \texttt{low} for both the target and optimizer models.
Following SkillOpt~\citep{yang2026skillopt}, we process the rollout minibatch in chunks when generating textual gradients and retrieval queries (Eq.~\ref{eq:query_update}).
The trajectories in each minibatch are split into chunks of size 8, and $\mathcal{A}_{\mathrm{query\_update}}$ produces gradient--query pairs for each chunk independently.
A single LLM call then merges the chunk-level outputs into the final set $\{(q_i,\delta_i)\}_{i=1}^{N}$, consolidating pairs that describe the same failure mode and removing redundant queries.
This design keeps each call within the context limit while allowing the gradients to reflect failure patterns across the entire minibatch.
Unless otherwise specified, we retrieve the top-$K=5$ sections per query.

Since the external skill corpus is collected from public repositories, it may contain skills written specifically for our evaluation benchmarks. 
We therefore construct a per-benchmark blocklist and remove it from the corpus before building the retrieval index.
For each benchmark, we search the corpus for keywords that identify it. 
The benchmark name and the names of the datasets it is built on. 
Whenever any skill matches, we exclude the entire repository containing it rather than the matched skill alone. 
We choose repository-level exclusion to be conservative. 
A repository that targets a benchmark often contains companion skills that encode benchmark-specific procedures without naming the benchmark, and such skills would evade keyword matching.

For each benchmark, we generate a task description $T$ and a harness description $H$ once, before any optimization run, and keep both fixed across all methods, models, and seeds.
Both descriptions are produced by a coding agent (codex) that reads the benchmark's harness code and states the task objective and output format ($T$) and the tools, file access, observation format, interaction budget, and scoring rule exposed by the harness ($H$).
The coding agent was not given access to test instances, and neither description was revised after observing test performance.

To isolate the contribution of retrieval, Retrieval-Free Skill Initialization (RFSI) follows exactly the same pipeline as RASI except for retrieval and adaptation.
Specifically, RFSI receives the identical $T$ and $H$, uses the same procedure and requirement decomposition $\{c_i\}_{i=1}^{M}$, and synthesizes the skill with the same skill-initialization system prompt; the only difference is that the grounded lessons $\mathcal{L}_{\text{init}}$ are omitted.
Hence, the gap between RASI and RFSI reflects the retrieved and adapted knowledge rather than the procedure decomposition or the prompt.
All skill update baselines start from this RFSI skill, so every update method is compared under the same $T$, $H$, and initial skill.

\subsection{Baseline Details}

\paragraph{Skill initialization.}
We compare RASI against three initialization strategies, all evaluated at zero rollouts.
\textbf{No skill} runs the agent without an initialized skill.
\textbf{SkillRouter}~\citep{zheng2026skillrouter} selects skills from a large library with a two-stage retrieve-and-rerank pipeline, in which a bi-encoder retrieves candidate skills and a cross-encoder reranks them using the full skill body.
We use it to retrieve a relevant skill from the external skill corpus.
\textbf{Retrieval-Free Skill Initialization (RFSI)} generates an initial skill with the target LLM directly from the task and harness descriptions, without access to the external skill corpus.

\paragraph{Skill update.}
We compare RASU against four update methods.
\textbf{TextGrad}~\citep{yuksekgonul2024textgrad} treats the skill as a text variable and optimizes it by backpropagating natural-language feedback, in which an LLM critiques execution results to produce textual gradients that are then used to revise the skill.
\textbf{GEPA}~\citep{agrawal2026gepa} evolves the skill through reflective mutation, in which an LLM reflects on execution traces to propose revisions, and maintains a Pareto front of candidates to preserve diverse strategies during selection.
\textbf{SkillOpt}~\citep{yang2026skillopt} converts scored rollouts into bounded add, delete, and replace edits on a skill document and accepts an edit only when it improves a held-out validation score.
\textbf{WikiSkill}~\citep{tang2026wikiskill} co-evolves skills with a persistent wiki, in which agent trajectories are consolidated into pages documenting failure modes and successful strategies, and skill updates are proposed by consulting this wiki.
Unless otherwise specified, we use the same model as both the target and optimizer model for all methods, and none of the baselines accesses the external skill corpus during optimization.

\subsection{Benchmark Details}

We evaluate on four benchmarks that cover diverse interaction settings, ranging from document-grounded question answering to spreadsheet manipulation, embodied household tasks, and web navigation.
For each benchmark, we use disjoint training, validation, and test splits ($\mathcal{D}_{\text{train}}$, $\mathcal{D}_{\text{val}}$, $\mathcal{D}_{\text{test}}$), whose sizes are summarized in Table~\ref{tab:benchmark-splits}.
Following SkillOpt~\citep{yang2026skillopt}, we use the same train/validation/test splits for all methods, so update acceptance and skill selection are performed on an identical validation set.

\paragraph{OfficeQA.}
OfficeQA~\citep{opsahl2026officeqa} is a benchmark for end-to-end grounded reasoning over enterprise documents.
Questions are posed over a corpus of U.S.\ Treasury Bulletins spanning nearly 100 years and about 89,000 pages, and require the agent to parse documents, retrieve relevant information across both unstructured text and tables, and perform numerical reasoning.
Answers are numerical and are scored as correct when they fall within an allowed relative error tolerance.
We use all 246 questions and split them into 50, 24, and 172 questions for training, validation, and testing, respectively.

\paragraph{SpreadsheetBench.}
SpreadsheetBench~\citep{ma2024spreadsheetbench} is a benchmark for real-world spreadsheet manipulation, built from questions collected from online Excel forums.
Each instruction requires the agent to read and modify spreadsheet files, and is evaluated in an online-judge style, where a solution is considered correct only if it produces the expected result on multiple spreadsheet files serving as test cases.
We use 400 tasks, split into 80, 40, and 280 tasks for training, validation, and testing, respectively.

\paragraph{ALFWorld.}
ALFWorld~\citep{shridhar2021alfworld} is a text-based embodied environment that aligns the ALFRED household tasks with the TextWorld engine.
The agent navigates a simulated household and interacts with objects through textual actions to complete goals from six task types: pick and place, examine in light, clean and place, heat and place, cool and place, and pick two and place.
We use 39 and 18 games for training and validation, and evaluate on the 134 out-of-distribution evaluation games.

\paragraph{WebShop.}
WebShop~\citep{yao2022webshop} is a simulated e-commerce environment containing about 1.18 million real-world products and 12,087 crowd-sourced instructions.
Given a natural-language instruction describing a desired product, the agent searches, navigates product pages, and selects options to make a purchase, and is rewarded according to how well the purchased product matches the requested attributes, options, and price.
We use 80 and 40 instructions for training and validation, and evaluate on the 500 test instructions.

\begin{table}[t]
\centering
\caption{Number of tasks in each split.}
\label{tab:benchmark-splits}
\vspace{6pt}
\small
\begin{tabular}{lccc}
\toprule
\textbf{Benchmark} & \textbf{Train} & \textbf{Validation} & \textbf{Test} \\
\midrule
OfficeQA          & 50 & 24 & 172 \\
SpreadsheetBench  & 80 & 40 & 280 \\
ALFWorld          & 39 & 18 & 134 \\
WebShop           & 80 & 40 & 500 \\
\bottomrule
\end{tabular}
\end{table}

\subsection{Skill Update Methods under Fixed Initialization}
\begin{table}[ht]
\begingroup
\definecolor{oursbg}{RGB}{234,242,248}
\setlength{\tabcolsep}{5.5pt}
\renewcommand{\arraystretch}{1}
\noindent
\begin{minipage}[t]{0.49\textwidth}\vspace{0pt}%
\centering
\caption{Update methods under a fixed RASI initialization with Qwen-3.5-9B. The highlighted row is RASO.}
\label{tab:opt-from-rasi-qwen}
\vspace{6pt}
\small
\begin{tabular}{llcc}
\toprule
\multicolumn{2}{c}{\textbf{Method}} & \multicolumn{2}{c}{\textbf{Benchmark}} \\
\cmidrule(lr){1-2}\cmidrule(lr){3-4}
\textbf{Init.} & \textbf{Update} & \textbf{OfficeQA} & \textbf{Spreadsheet} \\
\midrule
 & $-$       & 40.50\pmse{0.97} & 30.48\pmse{0.78} \\
 & TextGrad  & 38.76\pmse{1.40} & 30.48\pmse{0.78} \\
 & GEPA      & 38.18\pmse{0.70} & 31.07\pmse{0.21} \\
 & SkillOpt  & 40.70\pmse{0.89} & 30.60\pmse{0.66} \\
 & WikiSkill & 40.12\pmse{0.34} & 30.48\pmse{0.78} \\
\rowcolor{oursbg} \cellcolor{white}\multirow{-6}{*}{\textbf{RASI}}
 & \textbf{RASU} & \textbf{42.25}\pmseb{1.52} & \textbf{31.55}\pmseb{0.31} \\
\bottomrule
\end{tabular}
\end{minipage}\hfill
\begin{minipage}[t]{0.49\textwidth}\vspace{0pt}%
\centering
\caption{Update methods under a fixed RFSI initialization with Qwen-3.5-9B. The highlighted row is RASU.}
\label{tab:opt-from-rfsi-qwen}
\vspace{6pt}
\small
\begin{tabular}{llcc}
\toprule
\multicolumn{2}{c}{\textbf{Method}} & \multicolumn{2}{c}{\textbf{Benchmark}} \\
\cmidrule(lr){1-2}\cmidrule(lr){3-4}
\textbf{Init.} & \textbf{Update} & \textbf{OfficeQA} & \textbf{Spreadsheet} \\
\midrule
 & $-$       & 34.89\pmse{2.93} & 27.74\pmse{2.56} \\
 & TextGrad  & 36.82\pmse{1.59} & 25.95\pmse{0.63} \\
 & GEPA      & 36.43\pmse{1.97} & 24.05\pmse{1.56} \\
 & SkillOpt  & 37.21\pmse{1.21} & 29.52\pmse{2.21} \\
 & WikiSkill & 37.40\pmse{3.03} & 29.76\pmse{1.80} \\
\rowcolor{oursbg} \cellcolor{white}\multirow{-6}{*}{RFSI}
 & \textbf{RASU} & \textbf{38.48}\pmseb{1.25} & \textbf{30.60}\pmseb{2.82} \\
\bottomrule
\end{tabular}
\end{minipage}%
\endgroup
\end{table}

\begin{table}[ht]
\centering
\caption{Update methods under a fixed RFSI initialization with GPT-5.6-Luna. The highlighted row is RASU.}
\label{tab:opt-from-rfsi}
\vspace{6pt}
\begingroup
\definecolor{oursbg}{RGB}{234,242,248}
\setlength{\tabcolsep}{6pt}
\renewcommand{\arraystretch}{1}
\small
\begin{tabular}{llcc}
\toprule
\multicolumn{2}{c}{\textbf{Method}} & \multicolumn{2}{c}{\textbf{Benchmark}} \\
\cmidrule(lr){1-2}\cmidrule(lr){3-4}
\textbf{Initialization} & \textbf{Update} & \textbf{OfficeQA} & \textbf{Spreadsheet} \\
\midrule
 & $-$       & 40.11\pmse{0.58} & 44.40\pmse{2.74} \\
 & TextGrad  & 43.80\pmse{1.27} & 49.64\pmse{3.09} \\
 & GEPA      & 43.99\pmse{1.40} & 54.53\pmse{1.46} \\
 & SkillOpt  & 45.54\pmse{2.05} & 57.02\pmse{3.60} \\
 & WikiSkill & 41.47\pmse{1.18} & 47.14\pmse{3.98} \\
\rowcolor{oursbg} \cellcolor{white}\multirow{-6}{*}{RFSI}
 & \textbf{RASU} & \textbf{47.56}\pmseb{1.85} & \textbf{61.07}\pmseb{2.15} \\
\bottomrule
\end{tabular}
\endgroup
\end{table}
To isolate the contribution of the skill update procedure, we compare skill update methods while fixing the initial skill.
Table~\ref{tab:opt-from-rasi} in the main paper reports skill update results of GPT-5.6-Luna initialized with RASI.
Tables~\ref{tab:opt-from-rasi-qwen}, \ref{tab:opt-from-rfsi-qwen}, \ref{tab:opt-from-rfsi} report the remaining three combinations of initialization and LLM(Qwen-3.5-9B and GPT-5.6-Luna) on OfficeQA and SpreadsheetBench.
In all four combinations, RASU achieves the highest performance, showing that retrieval-augmented update improves over retrieval-free updates regardless of the initial skill and the target model.

\paragraph{Under LLM-init (RFSI).} 
With GPT-5.6-Luna (Table~\ref{tab:opt-from-rfsi}), RASU improves the initial skill by +7.45 and +16.67 points on OfficeQA and SpreadsheetBench, respectively, and outperforms the strongest competing baseline, SkillOpt, by +2.02 (47.56 vs.\ 45.54) and +4.05 (61.07 vs.\ 57.02). 
With Qwen-3.5-9B (Table~\ref{tab:opt-from-rfsi-qwen}), RASU improves the initial skill by +3.59 and +2.86 points and outperforms the strongest competing baseline, WikiSkill, by +1.08 (38.48 vs.\ 37.40) and +0.84 (30.60 vs.\ 29.76). 
In contrast, TextGrad and GEPA degrade the initial skill on SpreadsheetBench (25.95 and 24.05 vs.\ 27.74).

\paragraph{Under RASI.} 
Table~\ref{tab:opt-from-rasi-qwen} shows retrieval-free baselines struggle to improve upon the retrieval-augmented initial skill with Qwen-3.5-9B under RASI.
On OfficeQA, TextGrad, GEPA, and WikiSkill fall below the initial skill (38.76, 38.18, and 40.12 vs.\ 40.50), and SkillOpt improves it by only +0.20. 
On SpreadsheetBench, the largest gain among them is +0.59 (GEPA). 
RASU, in contrast, improves the initial skill by +1.75 and +1.07 points, reaching 42.25 and 31.55. 
TextGrad and WikiSkill return the initial skill unchanged on SpreadsheetBench, since no candidate update was improved over the validation set.

\paragraph{Complementarity of the two stages.} 
The two stages contribute gains that neither achieves alone. 
With Qwen-3.5-9B on OfficeQA, RASI without any update (40.50) already outperforms LLM-init (RFSI) followed by RASU (38.48), and combining both stages yields the best performance (42.25).

\subsection{Optimization Cost}
\begin{table}[t]
\centering
\caption{API cost (USD) of skill optimization with GPT-5.6-Luna. The lowest cost in each column is in bold.}
\label{tab:api-cost}
\vspace{6pt}
\begingroup
\definecolor{oursbg}{RGB}{234,242,248}
\setlength{\tabcolsep}{8pt}
\renewcommand{\arraystretch}{1}
\small
\begin{tabular}{lcccc}
\toprule
\textbf{Method} & \textbf{OfficeQA} & \textbf{Spreadsheet} & \textbf{ALFWorld} & \textbf{WebShop} \\
\midrule
TextGrad  & 1.45 & 0.97 & 2.04 & 41.03 \\
GEPA      & 1.69 & 0.58 & 1.79 & 18.05 \\
SkillOpt  & 2.23 & 2.24 & 2.75 & 17.86 \\
WikiSkill & 1.06 & \textbf{0.43} & 1.30 & 11.15 \\
\rowcolor{oursbg} \textbf{RASO} & \textbf{0.84} & 0.49 & \textbf{0.79} & \textbf{10.28} \\
\bottomrule
\end{tabular}
\endgroup
\end{table}

\begin{table}[t]
\centering
\caption{Number of rollouts used for skill optimization, excluding test-set evaluation.}
\label{tab:num-rollouts}
\vspace{6pt}
\begingroup
\definecolor{oursbg}{RGB}{234,242,248}
\setlength{\tabcolsep}{8pt}
\renewcommand{\arraystretch}{1}
\small
\begin{tabular}{lcccc}
\toprule
\textbf{Method} & \textbf{OfficeQA} & \textbf{Spreadsheet} & \textbf{ALFWorld} & \textbf{WebShop} \\
\midrule
TextGrad  & 196 & 320 & 114 & 360 \\
GEPA      & 272 & 360 & 174 & 370 \\
SkillOpt  & 236 & 360 & 154 & 400 \\
WikiSkill & 196 & 320 & 114 & 360 \\
\rowcolor{oursbg} \textbf{RASO} & 196 & 320 & 114 & 360 \\
\bottomrule
\end{tabular}
\endgroup
\end{table}

Tables~\ref{tab:api-cost} and~\ref{tab:num-rollouts} report the API cost and rollout count of skill optimization with GPT-5.6-Luna. 
All methods are optimized for the same 2 epochs. 
The rollout counts differ only because of how each method schedules evaluation. 
TextGrad, WikiSkill, SkillOpt and RASO update the skill from training rollouts and evaluate the updated skill on the validation set directly. 
On the other hand, GEPA first re-evaluates each updated skill on a support subset of the training set and proceeds to validation only when this yields an improvement, which incurs additional rollouts. 
SkillOpt performs an additional slow, meta update at the end of every epoch, which likewise adds rollouts. 
RASO therefore uses exactly as many rollouts as TextGrad and WikiSkill, and fewer than GEPA and SkillOpt. 

Under this budget, RASO achieves the lowest cost on three of the four benchmarks and remains within \$0.06 of the cheapest method on SpreadsheetBench. 
Since the budgets are matched, the savings reflect the cost of each rollout and optimization LLM call rather than the number of rollout. 
At an identical budget, RASO reduces the cost of TextGrad by 42--75\%, with the largest gap on WebShop (\$41.03 vs.\ \$10.28).

\newpage

\end{document}